\documentclass[runningheads]{llncs}
\usepackage{graphicx}
\begin{document}
\title{Hidden Markov Models and their Application for Predicting Failure Events}
\author{Paul Hofmann\inst{1} \and Zaid Tashman\inst{2}}
\authorrunning{Hofmann et al.}
\institute{Los Gatos, CA 95033, USA \email{paul@paul.email} \and San Francisco CA 94118, USA \email{zaid.tashman@accenture.com}}
\maketitle              % typeset the header of the contribution
\begin{abstract}
We show how Markov mixed membership models (MMMM) can be used to predict the degradation of assets. We model the degradation path of individual assets, to predict overall failure rates. Instead of a separate distribution for each hidden state, we use hierarchical mixtures of distributions in the exponential family. In our approach the observation distribution of the states is a finite mixture distribution of a small set of (simpler) distributions shared across all states. Using tied-mixture observation distributions offers several advantages. The mixtures act as a regularization for typically very sparse problems, and they reduce the computational effort for the learning algorithm since there are fewer distributions to be found. Using shared mixtures enables sharing of statistical strength between the Markov states and thus transfer learning. We determine for individual assets the trade-off between the risk of failure and extended operating hours by combining a MMMM with a partially observable Markov decision process (POMDP) to dynamically optimize the policy for when and how to maintain the asset.

\keywords{hidden Markov model \and Markov mixed membership model \and tied-mixture hidden Markov model \and HMM \and reinforcement learning \and POMDP \and partially observable Markov decision process \and time-series prediction \and asset degradation \and predictive maintenance}
\end{abstract}
\section{Introduction}

Predictive maintenance is an important topic in asset management. Up-time improvement, cost reduction, lifetime extension for aging assets and the reduction of safety, health, environment and quality risk are some reasons why asset intensive industries are experimenting with machine learning and AI based predictive maintenance.
 
Traditional approaches to predictive maintenance fall short in today’s data-intensive and IoT-enabled world \cite{ran2019survey}. In this paper we introduce a novel machine learning based approach for predicting the time of occurrence of rare events using Markov mixed membership  models (MMMM) \cite{fox2013mixed,murphy2012machine,zhang2015markov}. We show how we use these models to learn complex stochastic degradation patterns from data by introducing a terminal state that represents the failure state of the asset, whereas other states represent health-states of the asset as it progresses towards failure. The probability distribution of these non-terminal states and the transition probabilities between states are learned from non-stationary time-series data gathered as historic data, as well as real time streaming data (e.g. IoT sensors). 

Our approach is novel in two ways. First, we use an end-to-end  approach combining dynamic failure prediction of individual assets with optimization under uncertainty \cite{powell2016unified,powell2015tutorial} to find optimal replacement and repair policies. Typically, reinforcement learning approaches to predictive maintenance are satisfied with simple nondynamic prediction models \cite{koprinkova2013reinforcement}. Dynamic and more accurate failure prediction models are motivated by extending asset operating hours and are enabled by low cost cloud compute power. In section 5 we explain this in detail.

Secondly, we found several advantages using dynamic mixed membership modeling for remaining useful life estimates, over recurrent networks (specifically LSTM-based \cite{wu2018remaining}) and classical statistical approaches, like Cox-proportional hazard regression (CPHR) \cite{wei1989regression,satten1998inference}.

Adopting a Bayesian approach allows for starting with an estimate of the probability that can be subsequently refined by observation, as more sensor data is revealed in real time. In particular, our approach allows task specific knowledge to be included into the model. For example, the number of health-states, the number of mixtures of (topics, or archetypes) and the structure of the transition matrix may be modeled explicitly using engineering knowledge of practitioners. 

Typically, the data structure for LSTM is fixed, e.g. a matrix, or time-series, while MMMM is more flexible allowing different sampling frequencies for example. MMMM can also work with missing data out of the box, using expert knowledge as priors. A typical LSTM approach has to rely on transformation models using PCA for ad-hoc feature extraction for example, before being able to input time-series data into LSTM \cite{wu2018remaining}, thus separating the feature extraction part from the prediction part.

Further, LSTM-based approaches require complete episodic inputs to learn the prediction task, and thus can not be directly applied to right-censored
data. Right-censored data, or absorbing Markov chains, are needed for modeling degradation time-series unlike predicting classical time-series like for stock trends.

Traditionally, Cox-proportional hazard regression (CPHR) models with time-varying covariates are extensively used to represent stochastic failure processes \cite{wei1989regression,satten1998inference}. Though CPHR works well for right-censored data, it lacks the health-state representation of the asset. In other words, the proposed generative MMMM model can infer the probability of failure and the most likely health-state, whereas regression based models can only produce a probability estimate. This is particularly relevant in domains where the interpretability of the model results is important like in engineering. Further, the CPHR analysis allows only modeling relationships between covariates and the response variable, while MMMM enables modeling the relationships between any variable. That means, dynamic Bayseian networks and MMMM in particular, allow to model not only the relationships among covariates and the response variable, but allow to capture the relationships among the covariates too \cite{onisko2016interpret}. Understanding the full relationship between all covariates is important to understand asset degradation patterns.

The paper is structured as follows. Section 2 and 3 give a high level intro to HMM and HMM for failure prediction respectively. We are using the terminology of HMMs sharing hierarchical mixtures over their states; this being as a special case of MMMM. Section 4 explains how we use reinforcement learning. Section 5 is a tutorial explaining by example how this can be applied to a large scale system of hundreds of degrading assets.

\section{A brief introduction of the hidden Markov model}

A hidden Markov model is a generative graph model that represents probability distributions over sequences of observations \cite{ghahramani2001introduction}. It involves two interconnected models. The state model consists of a discrete-time, discrete-state first-order Markov chain $z_t$ $\in$ \{1,...,N\} that transitions according to $p(z_t|z_{t-1})$, while the observation model is governed by $p(x_t|z_t)$, where $x_t$ are the observations. The corresponding joint distribution of a sequence of states and observations can be factored as:

\begin{equation}
    p(z_{1:T},x_{1:T}) = p(z_1)p(x_1|z_1)\prod_{t=2}^T p(z_t|z_{t-1})p(x_t|z_t)
\end{equation}

Therefore, to fully define this probability distribution, we need to specify a probability distribution over the initial state $p(z_1)$, a $N \times N$ state transition matrix defining all transition probabilities $p(z_t|z_{t-1})$, and the emission probability model $p(x_t|z_t)$. To summarize, the HMM generative model has the following assumptions:

\begin{enumerate}
  \item Each observation $x_t$ is generated by a hidden state $z_t$.
  \item Transition probabilities between states $p(z_t|z_{t-1})$ represented by a transition matrix are constant.
  \item At time $t$, an observation $x_t$ has a certain probability distribution corresponding to possible hidden states.
  \item States are finite and satisfy first-order Markov property.
\end{enumerate}

\begin{figure}
    \centering
    \includegraphics[width=200px]{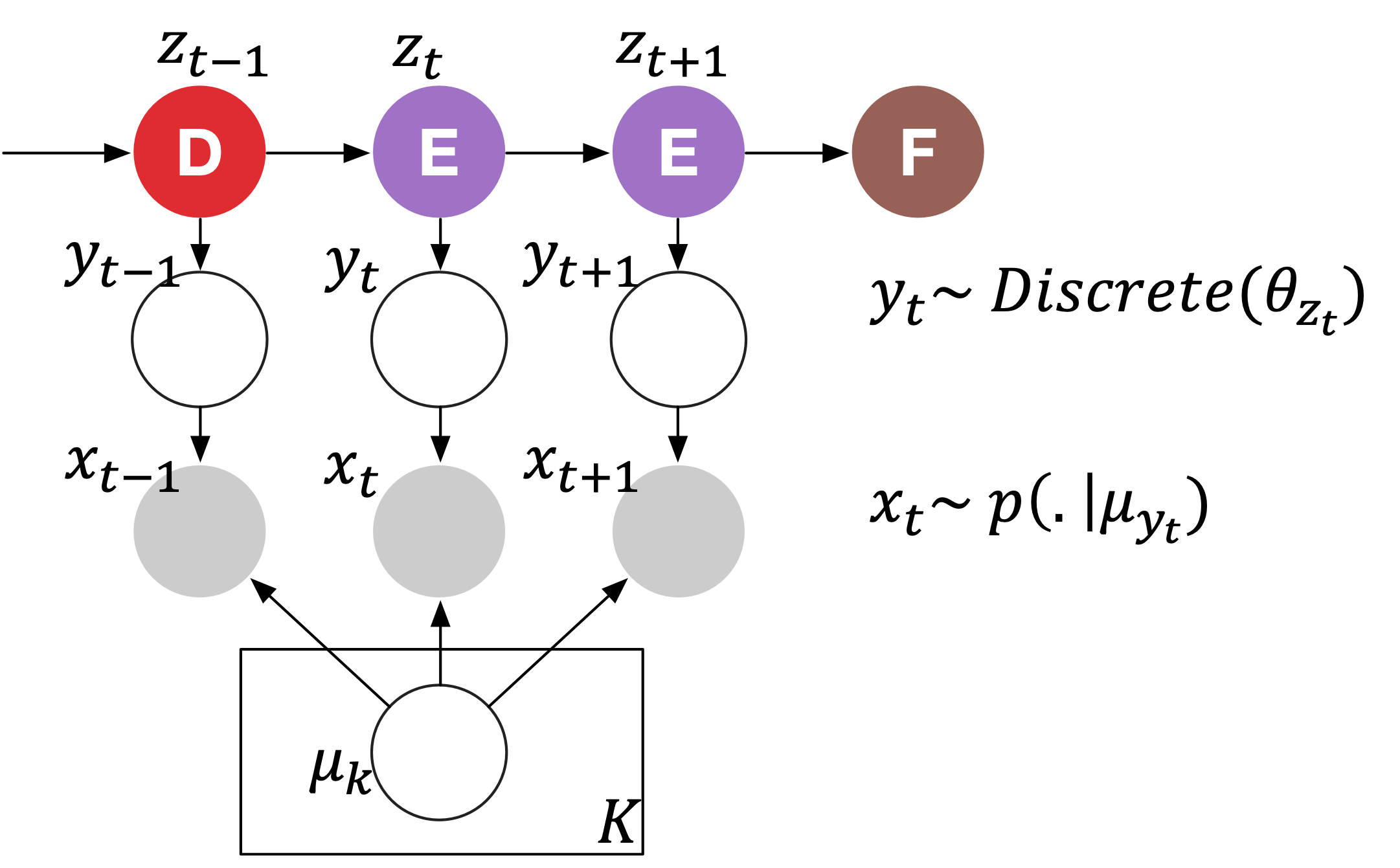}
    \caption{States share common distributions providing a way of regularizing the model to avoid overfitting}
    \label{fig:obs_model}
\end{figure}

The observation model specified by $p(x_t|z_t)$ can be represented by a discrete distribution (Bernoulli, Binomial, Poisson,...etc), a continuous distribution (Normal, Gamma, Weibull,...etc), or a joint distribution of many components assuming individual components are independent. In the work discussed in this paper we use a mixture distribution to represent the observation model. That is given a state $z_t$, the mixture component $y_t$ is drawn from a discrete distribution whose parameters $\theta$ are determined by the state $z_t$, denoted by $y_t\sim \emph{Discrete}(\theta_{z_t})$, where $\theta_{z_t}$ is the vector of mixing weights associated with state $z_t$. The observation $x_t$ is then drawn from one of a common set of $K$ distributions determined by component $y_t$, denoted as $x_t \sim p(.|\mu_{y_t})$, where $\mu_k$ is the parameters of the $k^{th}$ distribution. It is important to note that the mixture components $\mu$ are common and shared across states while the mixing weights $\theta$ vary across states. Coupling the mixture components across states provides a balanced trade-off between \emph{Heterogeneity} and \emph{Homogeneity} \cite{blei2003latent,blei2012probabilistic,teh2005sharing} allowing for information pooling across states, see Figure \ref{fig:obs_model}. This compromise is also beneficial when fitting HMM models with large number of states especially when certain states don't have enough observations (imbalanced data), as it provides a way of regularizing the model avoiding over-fitting. 

\subsection{Inference}

Once the parameters of a hidden Markov model distribution are learned from data, there are several relevant quantities that can be inferred from existing and newly observed data. For instance, given a partially observed data sequence $X = \{x_t, t=1,...,\tau\}$, what is the posterior distribution over the hidden states $p(z_t|x_{1:\tau})$ up to time $\tau < T$, the end of the sequence. This is a filtering task and can be carried out using the \emph{forward} algorithm. This posterior distribution will enable us to uncover the hidden health-state of an asset as we observe data streaming in. Additionally, one can be interested in computing the most probable state sequence path, $z^*$, given the entire data sequence. This is the maximum a posteriori (MAP) estimate and can be computed through the \emph{viterbi} algorithm \cite{forney1973viterbi}. Readers can find more information about the \emph{forward, forward-backward, Viterbi, and Baum-Welch} algorithms in \cite{murphy2012machine}.

% = \underset{z_{1:T}}{\arg\max p(z_{1:T}|y_{1:T})}$, $\underset{x_t \in X_t}{\mathrm{argmax}}

\section{Hidden Markov model for failure time prediction}

The model will represent the data as a mixture of different stochastic degradation processes. Each degradation process, a hidden Markov model, is defined by an initial state probability distribution, a state transition matrix, and a data emission distribution. Each of the hidden Markov models will have a terminal state that represents the failure state of the factory equipment. Other states represent health-states of the equipment as it progresses towards failure and the probability distribution of these non-terminal states are learned from data as well as the transition probabilities between states. Forward probability calculations enable prediction of failure time distributions from historical and real time data. Note that the data rate and the Markov chain evolution are decoupled allowing for some observations to be missing. Domain knowledge about the degradation process is important. Therefore, expert knowledge of the failure mechanism is incorporated in the model by enforcing constraints on the structure of the transition matrix. For example, not allowing the state to transition from an “unhealthy” state to a “healthy” state can be incorporated by enforcing a zero probability in the entries of the transition matrix that represent the probability of transition from an “unhealthy" state to a "healthy" state (see Figure \ref{fig:transition-matrix} for an example of a transition matrix with zeros to the left of the diagonal representing an absorbing Markov chain). Enforcing constraints on the transition matrix also reduces the computational complexity during model training as well as when the model is running in production for online prediction and inference. An important property of data generated from a fleet of factory equipment is right censoring. In the context of failure data, right censoring means that the failure times are only known in a few cases because for the vast majority of the equipment the failure time is unknown. Only information up to the last time the equipment was operational is known. Right censored observations are handled in the model by conditioning on the possible states the equipment can be in at each point in time, i.e. all non-terminal states. Once the model parameters are estimated, the model is used for different inferential tasks. As new data streams in from the asset, the state belief is calculated online in real time or recursively, which gives an estimate of the most probable health-state the asset is in. This is a filtering operation, which applies Bayes rule in a sequential fashion.

\begin{figure}
    \centering
    \includegraphics[width=345px]{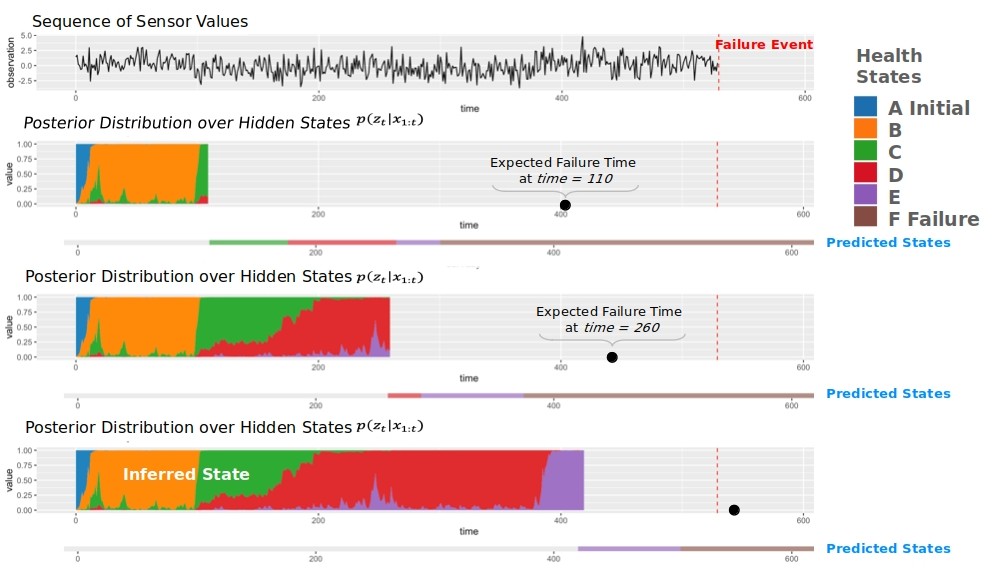}
    \caption{Failure Time Prediction using HMM}
    \label{fig:prediction}
\end{figure}

Another inference task is failure time prediction. As new data is streaming in, an estimate of the asset health-state over a certain future horizon is calculated as well as the most probable time at which the asset will enter the “failure” state (terminal state). Both of those inferential tasks are important as they provide a picture of the current state of the factory, as well as a forecast of when each asset will most likely fail; see Figure \ref{fig:prediction}. This information will then be used to optimize the decision-making process, to maintain or replace assets.

\section{Optimal decision making using partially observable Markov decision process}

At each time step we are confronted with a maintenance decision. Choosing the best action requires considering not only immediate effects but also long-term effects, which are not known in advance. Sometimes action with poor immediate effects can have better long-term ramifications. An "optimal" policy is a policy that makes the right trade-off between immediate effects and future rewards \cite{cassandra1994acting} and  \cite{kaelbling1998planning}. This is a dynamic problem due to the uncertainty of variables that are only revealed in the future. For example, sensors are not always placed in the right location on the equipment making the inference of the health-state of the asset noisy. There is also uncertainty about how the equipment will evolve over time, or how operators will use it. 

The goal of the optimal policy is to determine the best maintenance action to take for each asset at any given point in time, given the uncertainty of current and future health-states. We derive the policy from a value function, which gives a numerical value for each possible maintenance action that can be taken at each time step. In other words, a policy is a function that maps a vector of probability estimates of the current health-state to an action that should be taken at that time step. There is no restriction on this value function, which can be represented by neural networks, multi-dimensional hyper planes, or decision trees. 

\begin{figure}
    \centering
    \includegraphics[width=200px]{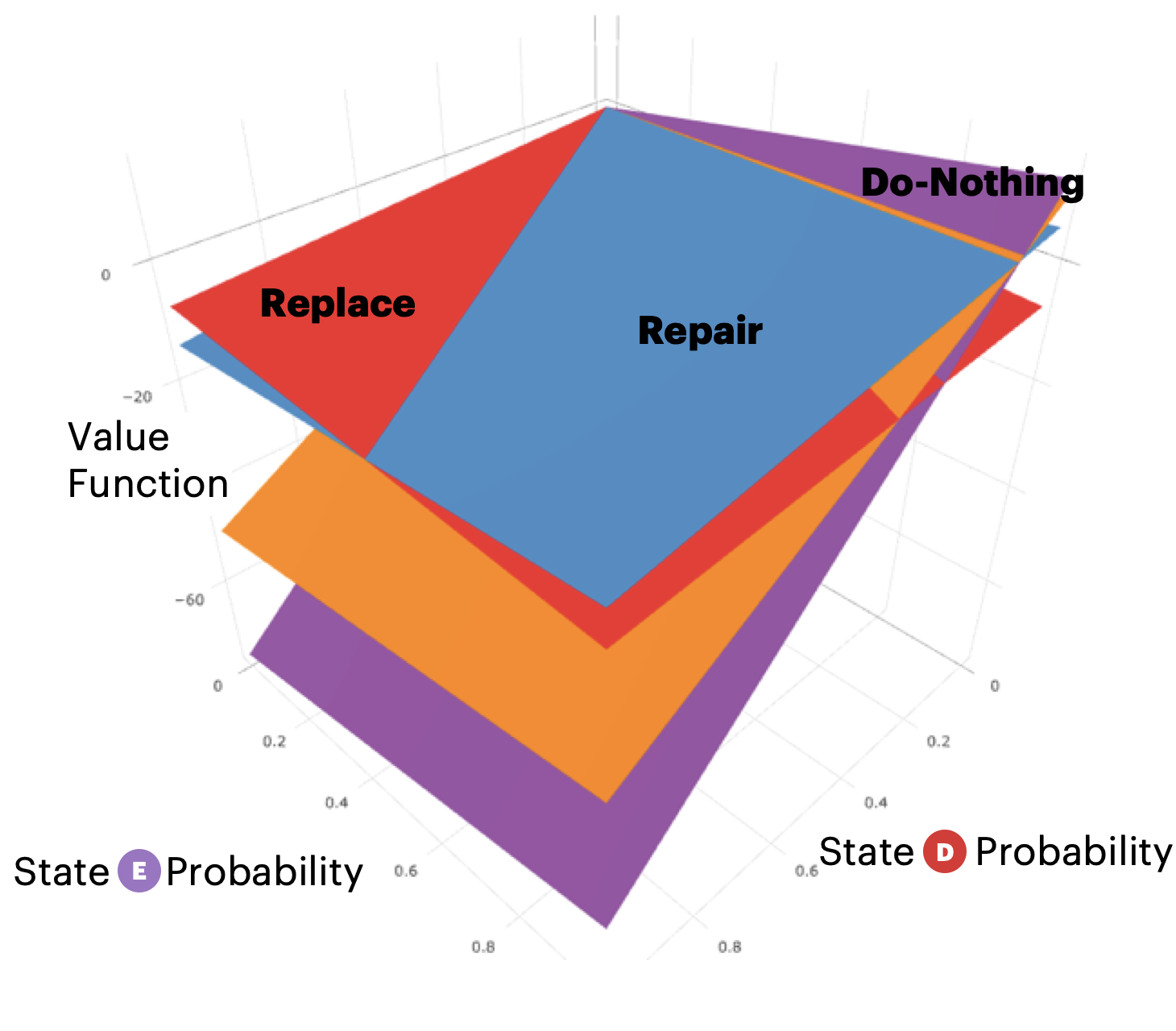}
    \caption{Value function}
    \label{fig:value-function}
\end{figure}

In this paper we focus on a local policy that is derived from a value function represented by multi-dimensional hyper planes. A hyper plane can be represented by a vector of its coefficients; therefore, the value function can be represented by a set of vectors; see Figure \ref{fig:value-function}.

In order to solve for our maintenance policy computing the value function, we assume that the model used for the degradation process of the asset is a hidden Markov model (HMM). Combining the dynamic optimization with the HMM enables us to use the parameters of our HMM to construct a partially observable Markov decision process (POMDP). A POMDP, in the context of asset modeling, is defined by:

\begin{equation}
\label{eq:pomdp}
POMDP = <S,A,T,R,X,O,\gamma,\pi>
\end{equation}

A set of health-states $S$, a set of maintenance actions $A$, an initial health-state belief vector $\pi$, a set of conditional transition probabilities $T$ between health-states, a cost or reward function $R$, a set of observations $X$, a set of conditional probabilities for the distribution of the observations $O$, and a discount factor $\gamma$ $\in$ [0,1]. Since our model of degradation is assumed to be a hidden Markov model, the states $S$, the transition probabilities $T$ and the initial probabilities $\pi$ of the POMDP are the same as the hidden Markov model parameters. The set of actions $A$ can be defined as $a_0$ = "Do Nothing", $a_1$ = "Repair", and $a_2$ = "Replace" for instance; see Figure \ref{fig:optimal-maintenance-policy} for example. This set is configurable based on the maintenance policy for the asset and how it is operated. Similar to $A$, the cost function $R$ is also configured based on maintenance policy and asset operation. The cost function $R$ typically includes the cost of failure, the cost of replacement, the cost of repair, and the negative cost of non-failure to name a few. In addition to financial cost, one can include other forms of cost like social cost of failure if the equipment failure could cause disruption to the environment for example, or cause shortage of supply. $R$ can be any type of function, production rules set up by the operator, look up tables, etc..

Once the POMDP is defined like in \ref{eq:pomdp}, we solve for the policy by computing the value function using a value iteration algorithm finding a sequence of intermediate value functions, each of which is derived from the previous one. The first iteration determines the value function for a time horizon of 1 time step, then the value function for a time horizon of 2 is computed from the horizon 1 value function, and so forth \cite{shani2013survey}.

Once the value function is computed, the best action to take on each asset at time t is determined by finding the action with the highest value given the current state probabilities at time t. 

\section{Examples}

There are many use cases for the methodology described. For example, the results of the failure prediction for a fleet of wind turbines may be input to reinforcement learning (POMDP) to optimize the schedule and route of the maintenance crews on a wind farm \cite{Hofmann:2017}. 

We illustrate our approach by going step by step through a real world example. We start with historic observations like multi-sensor time-series data for individual assets. We use expectation maximization (EM) to find the maximum likelihood or maximum a posterior (MAP) estimates of the parameters of our model from Fig \ref{fig:obs_model}. Typically, we use EM to solve larger problems with a few hundred sensors and more than 20 states. In our experience, EM is computationally more tractable for larger real world problems than Markov chain Monte Carlo (MCMC), or its modern variant, Hamilton Monte Carlo (HMC) \cite{hoffman2014no}), or the even more expensive variational inference approaches.

When using EM, the number of states for the HMM and the number of distinct distributions that make up our degradation processes are treated as hyper-parameters, which we input into our model. Posterior predictive checks (PPC) \cite{gelman2013bayesian} is then used to find the right set of hyper-parameters.

\begin{figure}
    \centering
    \includegraphics[width=345px]{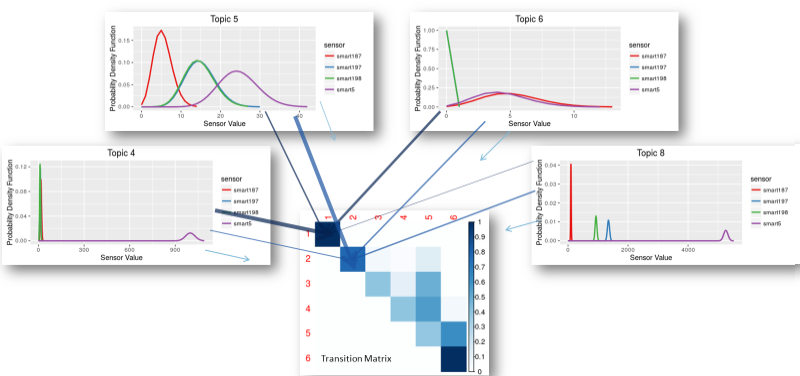}
    \caption{Example of transition matrix for absorbing MMMM}
    \label{fig:transition-matrix}
\end{figure}

In Figure \ref{fig:transition-matrix} we see an example of a transition matrix of a 6-state model. The off diagonal elements show the probabilities of transitioning between the HMM states, whereas the diagonals represent probabilities of remaining in each state. The lower triangular part of the matrix is set to zero to enforce an absorbing Markov chain where states move only from left to right towards the failure state. The Figure shows how the first two states are composed of mixtures, i.e. mixtures of common distributions. The thickness of the lines between the pdfs and the transition matrix indicates how much each common distribution contributes to the pertinent state distribution. For example, state 1 consists mainly of distribution 4 and 6, while distribution 5 contributes mainly to state 2. This shows an example of how the observation distributions of the states are mixtures of some set of common (simpler) distributions shared across all the states of the HMM. This approach can be seen as a generalization of tied-mixture HMM \cite{murphy2012machine}, where the shared distributions are limited to be  Gaussian, while we allow for hierarchical mixtures of all distributions of the exponential family. In our approach any topic, or archetype can a priori transition to any other archetype. Using a sparse Dirichlet prior on transition distributions we learn a meaningful dependence between archetypes
through posterior inference \cite{zhang2015markov}.

As mentioned before, expert knowledge of the failure mechanism maybe incorporated in the model by enforcing constraints on the structure of the transition matrix. 
 
 Figure \ref{fig:prediction} shows an example, of how we infer the hidden states given the sequence of observations, in our case, a time-series of sensor data. We see how the prediction of the expected failure time changes with additionally revealed sensor data over time. Using a Bayesian model like the MMMM enables us to calculate a new posterior for each newly observed data point thus gaining statistical strength and better prediction accuracy. Calculating the posterior is simple and short. It could be even done on edge devices. Contrast the simple solving of Bayes formula with the approach of discriminative models (i.e. regression models), where one would have to use the whole historic data set recalculating an improved model to add newly observed time series data.

\begin{figure}
    \centering
    \includegraphics[width=200px]{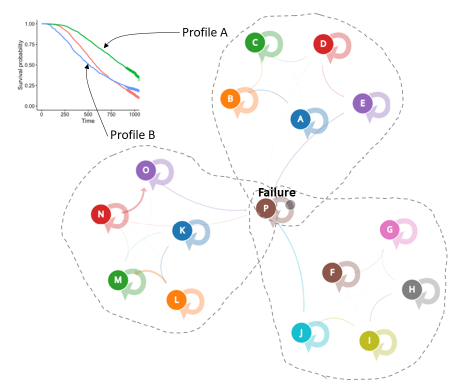}
    \caption{Degradation profiles share states}
    \label{fig:MMMM}
\end{figure}

Further, there is no obvious way using LSTM to get to a better prediction based on more real-time data points, since the posterior distribution is hidden in the weights. Simile to regression models, LSTM typically requires to run new back-propagation to learn from additional real time values. Another advantage of HMM of mixtures vs. LSTM is, it captures the natural (hidden) groupings within the data. Each group represents different asset profiles, i.e. distinct degradation processes, and thus failure curves (top left insert of Figure \ref{fig:MMMM}).

Figure \ref{fig:MMMM} shows an example of degradation states evolving over time. The thickness of the lines between the states indicates the probability of transitioning from one state to another. All assets start in state A, the initial state. After a certain amount of time, they end up in the terminal failure state P. See Figure \ref{fig:prediction} for an example how the states evolve over time (in this Figure the final state is called F). The other health-states (B to E, L to O, and F to J, for example) represent states of an asset as it progresses towards failure.

The data frequency and Markov chain evolution are decoupled allowing for real time data arriving at different rates. 

\begin{figure}
    \centering
    \includegraphics[width=345px]{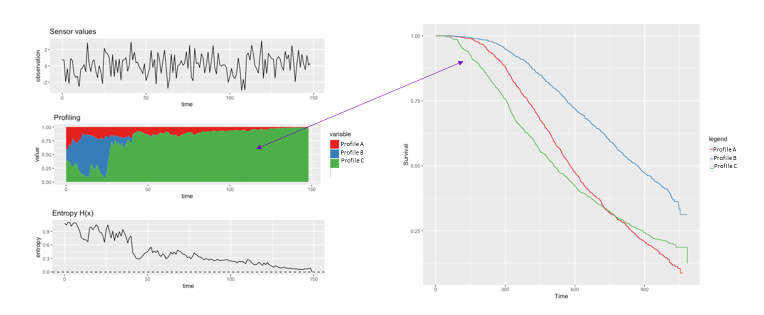}
    \caption{Profiles evolve as data is revealed leading to different survival curves}
    \label{fig:profiles-entropy}
\end{figure}

Once the model is fit, one can calculate the survival curves for the different degradation profiles which gives a summarized view of how assets fail as a baseline. See right plot in Figure \ref{fig:profiles-entropy}. The Figure also shows how the model can be used to "infer" which degradation profile the asset belongs to as new data arrives (colored graph on the left in the middle). The left bottom plot shows the entropy of the model's belief for a specific asset as more data is observed. Entropy, as a measure of uncertainty, decreases over time after more and more data points of the time-series have been reveled. The decreased entropy shows that after about 50 observations we can already be rather sure (entropy about 0.5) to which profile the asset at hand belongs. Thus the prediction for the profile and the life expectancy is rather reliable, after observing only a third of its life time (for this particular example of a degrading pump).

Having a measure of accuracy for failure prediction is very important for practitioners. Obviously, the traditional ROC curves are not a good choice since they do not capture the dynamic nature of our approach, i.e. recalculating new posteriors when new data points arrive. Typically, practitioners are facing trade-off questions. For example, what is the right point in time to replace a part. Replacing assets too early leads to unnecessary expenses. On average, parts are being replaced before they break. Running assets too long risks unforeseen down-time. To use such trade-offs as a measure for model quality is often more meaningful than ROC-type accuracy curves.

\begin{figure}
    \centering
    \includegraphics[width=300px]{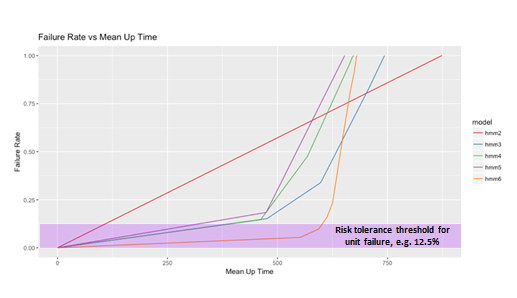}
    \caption{Model performance measured by complexity}
    \label{fig:complexity-critique}
\end{figure}

Risk tolerance is a typical constraint for operations managers. Using a trade-off diagram she can choose the model that predicts the longest operating hours given a certain risk level. Figure \ref{fig:complexity-critique} shows the trade-off between risk of failure vs. operating hours (mean up-time). Typically, the model that produces the fewest false negatives, i.e. the steepest hockey stick failure curve, is the most efficient. To the operator, the onset of the hockey  stick indicates the latest point in time for exchanging the asset, given a chosen risk level (12.5 percent in the pictured-example). Flat hockey stick failure curves, i.e. those with higher number of false positives, lead typically to reduced operating hours, since they indicate to the operator exchanging parts before their end of life-time. Sometimes, the steepest hokey stick curves come with less accuracy. The operator could mitigate reduced model accuracy by increasing spare parts inventory for example, thus still profiting from longer hours of asset operation. We see, ROC-type accuracy is not always the most important metric. The trade-off between failure rate and operating time can be more meaningful.

So far, we have shown in our example how we determine asset health evolving over time (Figure \ref{fig:prediction}). We are able to predict the degradation of individual assets by deriving profiles, which lead to different survival curves (Figure \ref{fig:profiles-entropy}).

Next, we have to derive actionable insights from the predictions by finding an optimal maintenance and resource allocation policy. We support the practitioner by determining the best action to take on an asset at any given moment, and assigning the right repair task to the right resource, i.e. who is to repair what and how.

\begin{figure}
    \centering
    \includegraphics[width=345px]{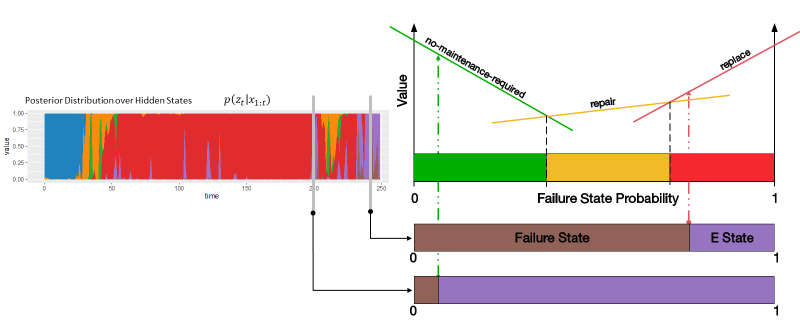}
    \caption{Optimal action changing over time}
    \label{fig:optimal-maintenance-policy}
\end{figure}

Before we can make decisions about repairing individual assets we need to understand which part of the value function (Figure \ref{fig:value-function}) to use, depending on a given asset health-state. Figure \ref{fig:optimal-maintenance-policy} shows how the best action changes over time depending on the transition state probabilities and the pertinent value function (green, yellow or red). States of assets are not observed directly but our model can be used to infer the posterior distribution over the hidden states. For example, the asset of Figure \ref{fig:optimal-maintenance-policy} has a low probability of failure around time point 200. According to the pertinent value function (policy) the best action is to "take no action". Around time point 230 the asset has high probability of being in a failure state (brown), thus, the value function recommends a "replace" action as the optimal take at this point in time.

\begin{figure}
    \centering
    \includegraphics[width=345px]{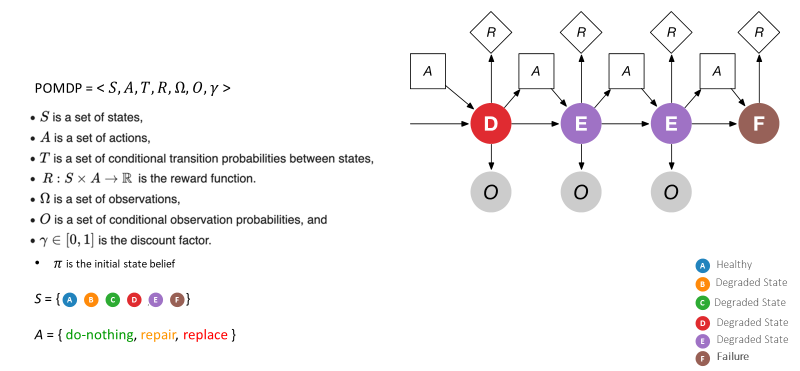}
    \caption{Selecting the best actions based on learned optimal policy
}
    \label{fig:POMDP}
\end{figure}

More quantitatively, not knowing the current (hidden) state we generalize a Markov decision process (MDP) to a partially observable MDP. Using POMDP we observe the state only indirectly relating it to the underlying hidden state probabilistically, Figure \ref{fig:optimal-maintenance-policy}. Being uncertain about the state of the asset, we introduce rewards $R$ (e.g. costs to repair, to replace, or the cost of down time), a set of actions $A$, and transition probabilities between health-states $T$; for details see equation \ref{eq:pomdp}. The transition probabilities $T$ and the initial probabilities $\pi$ of the POMDP are the same as our HMM parameters since we have used the hidden states $S$ of our HMM to model the degradation. The set of actions $A$ are $a_0$ = "Do Nothing", $a_1$ = "Repair", and $a_2$ = "Replace", see Figure \ref{fig:POMDP}.

We do not know the states, which are hidden. We only observe time dependent sensor data. From the observed sensor data we construct a belief, i.e. a posterior distribution over the states. From the belief we use the optimal policy, i.e. the solution of the POMDP, to find the optimal action to take, given the level of uncertainty. The POMDP solution is represented by a piecewise linear and convex value function calculated by the value iteration algorithm \cite{shani2013survey}. Once the value function is computed, the best action to take for each asset at time $t$ is determined by finding the action with the highest value given the current state probabilities at time $t$, as shown in Figure \ref{fig:POMDP}.

\section{Conclusions}

We showed how mixed membership hidden Markov models (MMMM) can be used to predict the degradation of assets. From historic observations and real time data, we modeled the degradation path of individual assets, as well as predicting overall failure rates. Using MMMM models has several advantages. Mixing over common shared distributions acts as a regularization for a typically very sparse problem, thus avoiding overfitting and reducing the computational effort for learning. Further, hierarchical mixtures (topics, or archetypes) allow for transfer learning through sharing of statistical strength between Markov states. We used a dual approach combining the MMMM failure prediction with a partially observable Markov decision process (POMDP) to optimize the policy for when and how to repair assets by determining the dynamic optimum between the risk of failure and extended operating hours of individual assets. We showed how to apply this approach using tutorial type of examples.

\bibliographystyle{splncs04}
\bibliography{references}

\end{document}